%% file: root.tex
\documentclass[letterpaper, 10 pt, conference]{ieeeconf}  

\IEEEoverridecommandlockouts                              

\usepackage{graphics} 
\usepackage{amsmath} 
\usepackage{amssymb}  
\usepackage{bm}
\graphicspath{{figures/}}

\usepackage{color}

\newcommand{\Real}{\mathbb{R}}
\newcommand\norm[1]{\left\lVert#1\right\rVert}

\usepackage{multirow}
\usepackage{graphicx}
\usepackage{url}

\title{\LARGE \bf
Humanoid Self-Collision Avoidance Using Whole-Body Control with Control Barrier Functions}

\author{Charles Khazoom$^{1*}$, Daniel Gonzalez-Diaz$^{1*}$, Yanran Ding$^1$, and Sangbae Kim$^{1}$
\thanks{*These authors have contributed equally to this work.}
\thanks{This work was supported by Disney Research Imagineering, Naver Labs and NSERC}
\thanks{$^{1}$Department of Mechanical Engineering Department,
        Massachusetts Institute of Technology,  77 Massachusetts Ave, Cambridge, MA, United States
        {\tt\small ckhaz@mit.edu, dgdiaz@mit.edu, yanran@mit.edu}}%
}

\begin{document}

\maketitle
\thispagestyle{empty}
\pagestyle{empty}

\begin{abstract}
This work combines control barrier functions (CBFs) with a whole-body controller to enable self-collision avoidance for the MIT Humanoid. Existing reactive controllers for self-collision avoidance cannot guarantee collision-free trajectories as they do not leverage the robot's full dynamics, thus compromising kinematic feasibility.
In comparison, the proposed CBF-WBC controller can reason about the robot's underactuated dynamics in real-time to guarantee collision-free motions.
The effectiveness of this approach is validated in simulation. First, a simple hand-reaching experiment shows that the CBF-WBC enables the robot's hand to deviate from an infeasible reference trajectory to avoid self-collisions.
Second, the CBF-WBC is combined with a linear model predictive controller (LMPC) designed for dynamic locomotion, and the CBF-WBC is used to track the LMPC predictions.
Walking experiments show that adding CBFs avoids leg self-collisions when the footstep location or swing trajectory provided by the high-level planner are infeasible for the real robot, and generates feasible arm motions that improve disturbance recovery.
\end{abstract}

\section{Introduction}
\input{Content/intro.tex}

\section{Control Barrier Functions}
\label{sec:CBF_bg}
\input{Content/CBF}

\section{Planning and Control Framework}
\label{sec:ctrl}
\input{Content/ControlFramework}


\section{Simulation Results and Discussion}
\label{sec:results}
\input{Content/Results}

\section{Conclusion}
\input{Content/Conclusion}

\addtolength{\textheight}{-12.5cm}   


\section*{Acknowledgments}
The authors wish to thank Matthew Chignoli for bringing up the potential of CBFs for humanoid self-collision avoidance.


\bibliographystyle{IEEEtran}
\bibliography{references/IEEEabrv,references/refs_humanoids_2022.bib,references/refs_YD.bib}

\end{document}

%% file: Content/intro.tex
\subsection{Motivation}
Humanoid robots already need to coordinate multiple joints to balance and locomote. Ensuring that the resulting stable motions do not damage the robot by colliding with itself adds an extra layer of complexity to the controller. Current approaches to address this challenge either use computationally expensive offline planners or real-time reactive strategies that do not leverage our knowledge about the robot's dynamics. Instead, these reactive approaches usually rely on carefully crafted potential functions and pose regularization to avoid self-collisions. However, the reactive method hinders the robot's performance and does not guarantee collision-free trajectories. To safely deploy humanoid robots in the real world, the robots should be able to simultaneously reason about their full-body dynamics and kinematics to ensure collision-free behaviors.
\par
This work combines a whole-body controller with control barrier functions, enabling the robot to reason about its full dynamics and effectively avoid self-collisions while achieving dynamic tasks.
\par
\subsection{Related Work}
Whole-body trajectory optimization can plan collision-free and dynamically feasible trajectories offline using non-linear optimization \cite{daiWholebodyMotionPlanning2014,schulmanMotionPlanningSequential2014,guReactiveLocomotionDecisionMaking2022}, but these methods are too slow to run in real-time for high-dimensional systems like humanoids.
\par
To enable real-time model predictive control for higher dimensional systems, reduced-order models are typically used \cite{dingDynamicWalkingFootstep2022,wensingHighspeedHumanoidRunning2013,sugiharaRealtimeHumanoidMotion2002}, but these models cannot reason about the full-body kinematics required to handle joint limits and self-collision constraints.
\par
Penalizing self-collisions using soft constraints has enabled real-time planning of collisions-free trajectories with a whole-body model predictive controller for a quadruped robot equipped with a robotic arm \cite{chiuCollisionFreeMPCWholeBody}. However, due to the use of soft constraints and to the limited number of solver iterations to satisfy real-time requirements, the trajectories may not always be collision-free.
\begin{figure}[!t]
\centering
\includegraphics[scale=1]{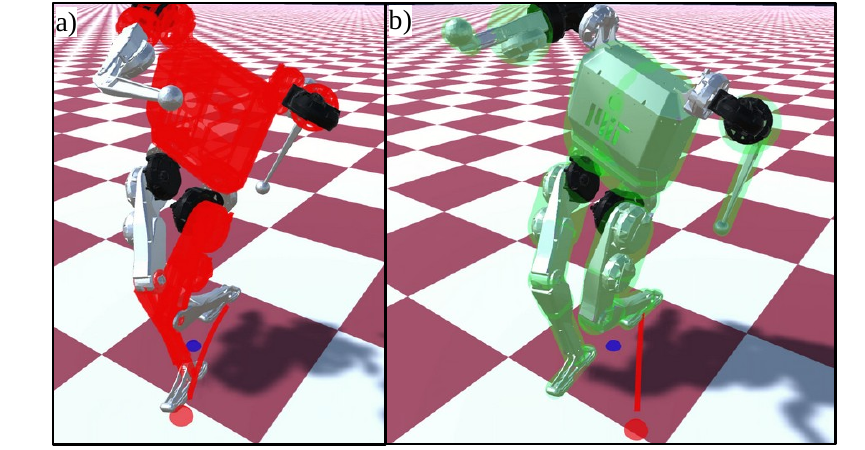}
\caption{Configuration of the MIT humanoid while walking after a lateral push. a) without control barrier functions, self-collisions occur. b) using control barrier functions within the whole-body controller avoids self-collisions. The green spheres and capsules approximate the robot's geometry for self-collision avoidance. The red bodies are in collision.}
\label{fig:with_without_collision}
\end{figure}
\par
Because of the limitations of planning approaches, joint limits and self-collisions are often avoided by reactive controllers that can run in real time. The most common approach is to use artificial potential fields (APFs) \cite{khatibRealtimeObstacleAvoidance1985} along with signed distance functions \cite{schwienbacherSelfcollisionAvoidanceAngular2011,dariushConstrainedResolvedAcceleration2010} to bias the robot towards feasible configurations. APFs are always active near the obstacle, even when moving away from it, which can lead to undesired behaviors. A more recent approach has used a learned self-collision boundary in joint space to solve for joint velocities that avoid self-collisions of a manipulator \cite{mirrazavisalehianUnifiedFrameworkCoordinated2018} and a humanoid robot \cite{koptevRealTimeSelfCollisionAvoidance2021}. Like APFs, this approach does not use prior knowledge of the robot's dynamics and thus cannot guarantee constraint satisfaction.
\par
Another strategy to encourage collision-free limb trajectories is to bias the robot towards a preferred nominal pose. This method has been shown to be particularly important when using centroidal momentum control, where unauthored upper-body joint trajectories are indirectly generated by commanding the robot's momentum \cite{schwienbacherSelfcollisionAvoidanceAngular2011, wensingImprovedComputationHumanoid2016, schullerOnlineCentroidalAngular2021}. Pose regularization can however limit the effectiveness of the upper limbs in rejecting perturbations because it limits the range of motion of the joints, thereby hindering emergent behaviors.
\par
Control barrier functions (CBFs) provide a general framework to guarantee safety of robotic systems, including legged robots \cite{grandiaMultiLayeredSafetyLegged2021,nguyen3DDynamicWalking2016}. Unlike artificial potential fields \cite{singletaryComparativeAnalysisControl2020}, CBFs leverage the non-linear dynamics of the system to constrain the robot's state to never leave a safe set. Typically, CBF constraints are used within a reactive safety filter on a nominal controller that can be formulated as a computationally efficient CBF-based quadratic program \cite{amesControlBarrierFunction2014a}.
\par
CBFs have previously been used for self-collision avoidance for fixed-based manipulators. Singletary \textit{et al.} used CBFs based on signed distance functions to successfully avoid self-collisions and collisions with the environment \cite{singletarySafetyCriticalManipulationCollisionFree2022}. The CBF constraints are enforced at the joint velocity level. As other approaches based on velocity-based inverse kinematics \cite{mirrazavisalehianUnifiedFrameworkCoordinated2018,koptevRealTimeSelfCollisionAvoidance2021,schwienbacherSelfcollisionAvoidanceAngular2011}, good or perfect velocity tracking is assumed in order to guarantee collision-free trajectories. However, velocity-based controllers cannot reason about the full dynamics of the robot. Therefore, they cannot benefit from desired feedforward acceleration and may require high feedback gains \cite{dariushConstrainedResolvedAcceleration2010}.
\par
For underactuated humanoid robots, acceleration-based whole-body controllers (WBC) are generally preferred to achieve dynamic locomotion because they can reason about the full dynamics of the robot to satisfy multiple tasks, and respect ground reaction force constraints \cite{wensingOptimizationControlDynamic2014,kimHighlyDynamicQuadruped2019}. Vector field inequalities \cite{vfi_Quiroz}, which can be equivalent to CBFs in some cases, have been used to impose collision avoidance constraints while reasoning about fixed-base dynamics. However, this approach has not been demonstrated for dynamic locomotion where multiple tasks need to be negotiated simultaneously while avoiding collisions and reasoning about underactuation. CBFs have previously been used in a whole-body controller for a quadruped robot \cite{grandiaMultiLayeredSafetyLegged2021}, but joint limits and self-collision were not considered. Given their importance for humanoid robots, enforcing joint limits and self-collision avoidance using CBFs for whole-body control has yet to be studied.
\par
\subsection{Contribution}
The primary contribution of this paper is the combination of CBFs with a whole-body controller for humanoid self-collision avoidance. First, the proposed CBF-WBC controller is shown to avoid self-collisions while tracking hand trajectories that would nominally collide. Second, the effectiveness of the approach is demonstrated for dynamic locomotion. The CBF-WBC is used to facilitate dynamic locomotion by tracking the references generated from a previously developed locomotion predictive controller (LMPC) \cite{dingDynamicWalkingFootstep2022}. Specifically, the CBFs enable collision-free leg trajectories and arm motions that aid in disturbance recovery.
\par
The remainder of the paper is organized as follows. Section \ref{sec:CBF_bg} presents a background on CBFs. Section \ref{sec:ctrl} details the LMPC, the humanoid model, the CBFs used for joint limit and self-collisions avoidance and the CBF-WBC. Section \ref{sec:results} demonstrates the benefits of the CBF-WBC on various scenarios.

%% file: Content/CBF.tex
The proposed CBF-WBC framework leverages control barrier functions \cite{amesControlBarrierFunctions2019} to avoid collisions. 
\par 
Consider a nonlinear control-affine system:
\begin{equation}
    \dot{\bm{x}} = \bm{f}(\bm{x}) + \bm{g}(\bm{x})\bm{u},
\end{equation}
where $\bm{x} \in \Real^n$ is the system state, $\bm{u} \in \Real^m$ is the control input. If a Lipschitz continuous control law $\bm{u} =\bm{k}(\bm{x},t)$ can be defined, then a unique trajectory can be guaranteed given an initial state $\bm{x}(0)$.

CBFs guarantee safety in the form of forward invariance on a safe set $\mathcal{C}$, defined as the zero superlevel set of a continuously differentiable scalar function $h(\bm{x}):\mathbb{R}^n\rightarrow\mathbb{R}$,
\begin{equation} 
    \mathcal{C} = \{~\bm{x} \in \Real^n~|~h(\bm{x}) \geq 0~\}.
\end{equation}

If the solution of $\bm{x}(t) \in \mathcal{C} $ $\forall~t$, then the system remains safe and the set $\mathcal{C}$ is forward invariant.

CBFs can be used to ensure the safety of a nominal control law $\bm{k}(\bm{x},t)$ that might not be necessarily safe. This can be done by bounding the time derivative of $h$ with the inequality \eqref{eq:cbf_constraint_general}
\begin{equation}
    \dot{h}(\bm{x},\bm{u}) = \frac{\partial h(\bm{x})}{\partial \bm{x}}
    [\bm{f}(\bm{x})+\bm{g}(\bm{x})\bm{u}] \geq -\gamma(h(\bm{x})),
    \label{eq:cbf_constraint_general}
\end{equation}
where $\gamma: \Real \rightarrow \Real$ belongs to an extended class of functions $\mathcal{K}_\infty$. In practice, this function is chosen as $\gamma(h)=\alpha_1 h$ for simplicity, with $\alpha_1 \in \Real_+$.
\par
Because the system is control-affine, the CBF constraint (\ref{eq:cbf_constraint_general}) can be enforced in the quadratic program (QP) \eqref{eq:CBF_filter} that minimally modifies the control input $\bm{k}(\bm{x},t)$ such that the system remains safe \cite{amesControlBarrierFunction2014a}:
\begin{align}
        \bm{u}^* = \arg\min_{\bm{u}} \quad & \frac{1}{2} \norm{\bm{u}-\bm{k}(\bm{x},t)} ~~~~~~\textrm{(CBF-QP)}\nonumber\\ 
        \text{subject to} \quad &  \dot{h}(\bm{x},\bm{u}) \geq -\alpha_1 h(\bm{x})\label{eq:CBF_filter}
\end{align}
\par
In a system with generalized coordinates $\bm{q} \in \Real^{n_q}$ and velocities $\dot{\bm{q}} \in \Real^{n_q}$ where the barrier function $h$ only depends on $\bm{q}$, its time derivative can be written as
\begin{equation}
    \dot{h}(\bm{q},\dot{\bm{q}})= \frac{\partial h(\bm{q})}{\partial \bm{q}} \dot{\bm{q}},
\end{equation}
and the consequent barrier constraint can be written as
\begin{equation}
    \dot{h}(\bm{q},\dot{\bm{q}}) \geq -\alpha_1 h(\bm{q})\label{eq:vel_CBF_constr}.
\end{equation}
Since the barrier constraint \eqref{eq:vel_CBF_constr} does not depend on the joint accelerations, it cannot be enforced directly for systems with relative degree 2, where torque inputs affects the second derivative of $\bm{q}$. Grandia \textit{et al.} \cite{grandiaMultiLayeredSafetyLegged2021} showed how exponential CBFs \cite{nguyenExponentialControlBarrier2016} can be used to circumvent this limitation by creating a new barrier function $h_e(\bm{q},\dot{\bm{q}})$:
\begin{equation}
    h_e(\bm{q},\dot{\bm{q}})\triangleq \dot{h}(\bm{q},\dot{\bm{q}}) + \alpha_1 h(\bm{q})\label{eq:eCBF_def},
\end{equation}
and enforcing the following constraint on the time derivative of $h_e$:
\begin{equation}
\dot{h}_e(\bm{q},\dot{\bm{q}},\Ddot{\bm{q}}) = \frac{\partial  h_e(\bm{q},\dot{\bm{q}})}{\partial \bm{q}} \dot{\bm{q}} + \frac{\partial  h_e(\bm{q},\dot{\bm{q}})}{\partial \dot{\bm{q}}} \Ddot{\bm{q}}  
    \geq -\alpha_2 h_e(\bm{q},\dot{\bm{q}})
    \label{eq:hd_e_constraint},
\end{equation}
with $\alpha_2\in \Real_+$.
The constraint \eqref{eq:hd_e_constraint} is linear in the joint accelerations $\ddot{\bm{q}}$ and can be introduced as a linear inequality into a QP-based whole-body controller that solves for joint torques and accelerations as explained in section \ref{sec:CBF-WBC}.

%% file: Content/ControlFramework.tex
This section describes the planning and control framework of the proposed control barrier function-based whole-body controller (CBF-WBC). First, the previously developed humanoid locomotion controller \cite{dingDynamicWalkingFootstep2022} using linear model predictive control (LMPC) is described as the high-level planner; then, the dynamics model of the MIT Humanoid and the proposed CBF-WBC are introduced.

\subsection{Linear Model Predictive Control}\label{sec:LMPC}

The locomotion controller tailored for the MIT Humanoid is used in this work as a high-level planner that produces centroidal reference trajectories and footstep locations \cite{dingDynamicWalkingFootstep2022}. The single rigid body (SRB) dynamics model is used, where the state vector $\bm{x}\in\mathbb{R}^{15}$ is augmented with the current foot location $\bm{c}\in\mathbb{R}^3$. The input includes the ground reaction wrench $\bm{u}\in\mathbb{R}^6$ and the stepping vector $\delta\bm{c}\in\mathbb{R}^3$.


The desired control is obtained by solving a finite horizon optimal control problem with a discrete-time linear dynamics
\begin{equation}\label{eq:discrete_dynamics}
	\bm{x}_{k+1} = \bm{A}\bm{x}_{k}+\bm{B}\bm{u}_{k}+\bm{d}+\eta_k\bm{A}_c\delta\bm{c},
\end{equation}
where the matrices $\bm{A},\bm{B},\bm{d}$ are obtained with forward Euler integration scheme. 

The last term in \eqref{eq:discrete_dynamics} allows LMPC to reason about stepping strategy in a unified optimization framework. Specifically, $\bm{A}_c$ is a sub-matrix that takes the last three columns of $\bm{A}$, and $\bm{A}_c \delta\bm{c}$ quantifies the effect of taking a step $\delta\bm{c}$ on the state evolution. Since the robot can only take a step at certain instances, a binary parameter $\eta_k$ encodes the gait schedule. $\eta_k=1$ signifies that the robot is expecting to take a step at time $k$, and $\eta=0$ otherwise.

\subsection{The Humanoid Model}
The MIT Humanoid is modeled with line feet and is actuated by 18 joints, including 5 degrees of freedom per leg and 4 degrees of freedom per arm. Its configuration is described by $\bm{q} = \left[\bm{q}_b^\top~\bm{q}_a^\top\right]^\top$, where $\bm{q}_b \in \Real^{6}$ and $\bm{q}_a \in \Real^{18}$ are the unactuated torso and actuated joints, respectively. The robot's full dynamics are described by
\begin{equation}
    \bm{H}(\bm{q})\ddot{\bm{q}} + \bm{C}(\bm{q},\dot{\bm{q}}) = \bm{S}_a\bm{\tau} + \sum_{c\in \{\mathrm{R,L}\}} \bm{J}_{c}^\top(\bm{q})\bm{F}_c,\label{eq:full_dyn}
\end{equation}
where $\bm{H}(\bm{q}) \in \Real^{24 \times 24}$ is the mass matrix, $\bm{C}(\bm{q},\dot{\bm{q}}) \in \Real^{24}$ is the Coriolis and gravitational term, $\bm{S}_a = \left[\bm{0}^{18\times6}~\bm{I}^{18\times18}\right]^{\top}$ is a matrix that selects the actuated joints torques $\bm{\tau} \in \Real^{18}$, $\bm{J}_{c}(\bm{q}) \in \Real^{6\times 5}$ is the line contact Jacobian of the foot $c \in \{\mathrm{R,L}\}$ contacting the ground and $\bm{F}_c\in \Real^{5}$ is the ground reaction wrench from the contacting line foot $c$.

\subsection{Whole-Body Control with Control Barrier Constraints}
\label{sec:CBF-WBC}
The LMPC uses a reduced-order model that captures the dominant dynamics of humanoid locomotion and provides reference footstep location and torso pose that account for the long-horizon effects. The CBF-WBC considers the full dynamics of the robot to track the task-space reference trajectories from LMPC while avoiding self-collisions and respecting joint limits with CBF inequality constraints.
\par
\subsubsection{Whole-Body Control Tasks}
The CBF-WBC negotiates a combination of various tasks depending on the desired behavior. These tasks include a torso pose task, a swing foot pose task, an arm end-effector position task to track desired hand positions, a joint position task to regularize the robot's pose towards a nominal configuration, and a centroidal angular momentum task that generates arm motions that assist walking.
\par
The commanded task acceleration $\dot{\bm{v}}_i^\mathrm{cmd}$ is composed of the feed-forward term $\ddot{\bm{x}}_i^\mathrm{des}$ and PD feedback terms with gains $\bm{K}_{p,i}$ and $\bm{K}_{d,i}$
\begin{equation}
    \dot{\bm{v}}_i^\mathrm{cmd} = \ddot{\bm{x}}_i^\mathrm{des}+\bm{K}_{p,i}(\bm{x}_i^\mathrm{des} - \bm{x}_i) + \bm{K}_{d,i} (\dot{\bm{x}}_i^\mathrm{des} - \dot{\bm{x}}_i),
\end{equation}
where $\bm{x}_i$ and $\bm{x}_i^\mathrm{des}$ are the measured and desired task position, respectively. For the centroidal angular momentum task, the robot's angular momentum is commanded as in \cite{wensingImprovedComputationHumanoid2016}.

\subsubsection{CBFs for Joint Limits}
\label{sec:CBF_JOINT_LIMIT}
The joint limits in the form of box constraints $q_j^\mathrm{lb} \le q_j \le q_j^\mathrm{ub}$ are imposed using CBFs. The corresponding safe sets are defined as $\mathcal{C}_j^\mathrm{lb}$ and $\mathcal{C}_j^\mathrm{ub}$ for the position $q_\mathrm{j}$ of the actuated joint $j \in \{1,...,18\}$
\begin{gather}
    \mathcal{C}_j^\mathrm{lb} = \{~q_{j} \in \Real~|~h_j^\mathrm{lb}(q_{j}) = q_{j} - q_j^\mathrm{lb} \geq 0~\}\label{eq:CBF_jmin}\\
    \mathcal{C}_j^\mathrm{ub} = \{~q_{j} \in \Real~|~h_j^\mathrm{ub}(q_{j}) = q_j^\mathrm{ub} - q_{j} \geq 0~\}\label{eq:CBF_jmax}
\end{gather}

Equations \eqref{eq:eCBF_def} and \eqref{eq:hd_e_constraint} are applied to the barrier functions \eqref{eq:CBF_jmin} and \eqref{eq:CBF_jmax} to obtain the inequalities
\begin{gather}
    \ddot{q}_j + (\alpha_{1,j} + \alpha_{2,j})\dot{q}_j + \alpha_{1,j}\alpha_{2,j}(q_j - q_j^\mathrm{lb}) \ge 0 \label{eq:CBF_ineq_jmin}\\
    -\ddot{q}_j - (\alpha_{1,j} + \alpha_{2,j})\dot{q}_j + \alpha_{1,j}\alpha_{2,j}(q_j^\mathrm{ub} - q_j) \ge 0 \label{eq:CBF_ineq_jmax},
\end{gather}
where $\alpha_{1,j} \in \Real_+$ and $\alpha_{2,j} \in \Real_+$  are positive scalars that are set equal for simplicity. 
The inequalities \eqref{eq:CBF_ineq_jmin} and \eqref{eq:CBF_ineq_jmax} are added in the CBF-WBC QP \eqref{eq:CBF-WBC-QP} to guarantee the forward invariance of the sets $\mathcal{C}_j^\mathrm{lb}$ and $\mathcal{C}_j^\mathrm{ub}$.

\subsubsection{CBFs for Self-Collisions}
\input{Content/ControlFramework_SelfCollisions}

\subsubsection{CBF-WBC Formulation}
The CBF-WBC is formulated as a quadratic program that solves for the joint accelerations $\ddot{\bm{q}}$, joint torques $\bm{\tau}$ and ground reaction wrench $\bm{F}_c$ given the robot's current state $\left[\bm{q}^\top~ \dot{\bm{q}}^\top\right]^\top$:
\begin{subequations}\label{eq:CBF-WBC-QP}
	\begin{align}
		\underset{\ddot{\bm{q}},\tau,\bm{F}_c}{\text{minimize}} ~~~~ 
		    &\sum_i\norm{\bm{J}_i\ddot{\bm{q}}+\dot{\bm{J}}_i \dot{\bm{q}} - \dot{\bm{v}}_i^\mathrm{cmd}}_{Q_i}\nonumber\\ 
		    & + \norm{\bm{F}_c - \bm{F}_c^\mathrm{cmd}}_{R_F} + \norm{\bm{\tau}}_{R_{\tau}}\nonumber\\
		\text{subject to} ~~~~&\mathrm{Robot~Full~Dynamics} ~~\eqref{eq:full_dyn}\nonumber\\
		    & \bm{J}_c\ddot{\bm{q}}+\dot{\bm{J}}_c\dot{\bm{q}}=\bm{0} \label{eq:line_contact_QP}\\
		    & \bm{F}_c \in \bm{\mathcal{F}} \label{eq:fr_cone_QP}\\
		   &\norm{\bm{\tau}} \leq \bm{\tau}_\mathrm{max}\label{eq:wbc_bnds}\\
		   &\mathrm{CBFs~for~Joint~Limits}~~ \eqref{eq:CBF_ineq_jmin},\eqref{eq:CBF_ineq_jmax}\nonumber\\
		   &\mathrm{CBFs~for~Self\textrm{-}Collisions}~~\eqref{eq:coll-CBF-ineq}\nonumber,
		   \label{eq:simple bounds}
	\end{align}
\end{subequations}
where $\bm{J}_i$ is the Jacobian for task $i$, $\dot{\bm{v}}_i^\mathrm{cmd}$ is the commanded task acceleration, $\bm{F}_c^\mathrm{cmd}$ is a commanded ground reaction force and $Q_i$, $Q_h$, $R_F$ and $R_{\tau}$ weigh each term in the cost function. The constraints respect the robot's full dynamics \eqref{eq:full_dyn}, the line contact constraint \eqref{eq:line_contact_QP} that enforces zero linear acceleration and zero pitch and yaw angular acceleration of the line foot, friction cone constraint \eqref{eq:fr_cone_QP} and torque limits $\bm{\tau}_\mathrm{max}$. Although \eqref{eq:wbc_bnds} could conflict with the CBF constraints and render the QP infeasible, this did not happen in practice due to the high torque density of the MIT humanoid \cite{chignoliMITHumanoidRobot2021}.
\par
A novel aspect of the CBF-WBC is the use of the CBF constraints \eqref{eq:CBF_ineq_jmin}, \eqref{eq:CBF_ineq_jmax} and \eqref{eq:coll-CBF-ineq} to avoid joint limit violations and self-collisions. Incorporating these CBF constraints into the CBF-WBC allows the robot to reason holistically about kinematic and dynamic feasibility while minimizing task error and respecting contact and friction cone constraints.

%% file: Content/ControlFramework_SelfCollisions.tex
Self-collision avoidance constraints are imposed as CBF constraints based on the signed distance between geometric primitives. The safe set is the set of joint configurations where there are no collisions between any two bodies. Without loss of generality, let $AB$ denote a collision pair of geometric primitives $A$ and $B$. The safe set for the pair $AB$ is defined as
\begin{equation}
    \mathcal{C}_{AB}^\mathrm{sd} = \{~\bm{q} \in \Real^{24}~|~h^\mathrm{sd}_{AB}(\bm{q}) \geq 0~\},
\end{equation}
where $h^\mathrm{sd}_{AB}(\bm{q})$ is the signed distance for collision pair $AB$, with $AB \in \{1,2,..., N^\mathrm{sd}_\mathrm{pairs}\}$.
\par
Spheres and capsules approximate the robot's geometry for efficient computation (Fig. \ref{fig:with_without_collision}b). Signed distance between spheres and capsules is computed by finding the minimum distance between points and line segments using the algorithm outlined in \cite{lumelskyFastComputationDistance1985} and subtracting the positive distance from the sum of their radii to check if the bodies are in penetration

\begin{equation}
    h^\mathrm{sd}_{AB}(\bm{q})= ||\bm{p}^r_A -\bm{p}^r_B ||_2 - (\rho_A+\rho_B)\label{eq:SDF-CBF}
\end{equation}
where $\bm{p}^r_A$ is the center of the sphere or the closest point along a line segment that defines a capsule and $\rho_A$ is the radius of collision body $A$.



The gradient of the signed distance function with respect to the joint positions of the robot is computed as in \cite{schulmanMotionPlanningSequential2014} and \cite{chiuCollisionFreeMPCWholeBody}.
\begin{equation}
    \bm{J}_{AB}(\bm{q}) =\frac{\partial h^\mathrm{sd}_{AB}(\bm{q})}{\partial \bm{q}}= \pm \hat{\bm{n}}^T (\bm{J}_A(\bm{q})-\bm{J}_B(\bm{q}))
\end{equation}
\begin{equation}
    \bm{J}_A(\bm{q}) = \frac{\partial \bm{FK}_{p_A}(\bm{q})}{\partial \bm{q}}
\end{equation}
\begin{equation}
    \bm{p}_A= \bm{p}^r_A + \rho_A \bm{\hat{n}}
\end{equation}
where $\bm{J}_{AB}:\Real^{24} \rightarrow \Real^{24}$ takes the sign of $h^\mathrm{sd}_{AB}$, $\hat{\bm{n}} \in \Real^{3}$ is the unit vector that is in the direction of the closest points on the bodies $A$ and $B$, $\bm{J}_A: \Real^{24} \rightarrow \Real^{3\times24}$ is the Jacobian of the forward kinematics of the robot $\bm{FK}_{p_A}: \Real^{24} \rightarrow \Real^3$ at the closest point $\bm{p}_A$.

To enforce the self-collision constraints at the acceleration level, as in section \ref{sec:CBF_JOINT_LIMIT}, we apply equations  \eqref{eq:eCBF_def} and \eqref{eq:hd_e_constraint} to the barrier function \eqref{eq:SDF-CBF} and add the following inequalities in the CBF-WBC
\begin{equation}
\bm{J}_{AB}\ddot{\bm{q}} + \dot{\bm{J}}_{AB} \dot{\bm{q}}+(\alpha_1^{AB}+\alpha_2^{AB})\bm{J}_{AB}\dot{\bm{q}} +\alpha_1^{AB} \alpha_2^{AB} h^\mathrm{sd}_{AB} \geq 0\nonumber
\end{equation}
\vspace{-7.2mm}
\begin{flalign}
\label{eq:coll-CBF-ineq}
&&\forall AB \in \{1,2,..., N^\mathrm{sd}_\mathrm{pairs}\},
\end{flalign}
where $\alpha_1^{AB}\in \Real_+$ and $\alpha_2^{AB}\in \Real_+$ are also chosen to be equal. The time derivative of $\hat{\bm{n}}$ is neglected when computing $\dot{\bm{J}}_{AB} \dot{\bm{q}}$ for simplicity.
\par
Although only spheres and cylinders are used, the approach can be extended to more complex geometries such as mesh representations since the signed distance gradient approximation only requires computing the distance and location of the closest points.

%% file: Content/Results.tex
This section first provides the implementation details for the simulated experiments. Next, these experiments and their corresponding results are described. These results are also presented in the accompanying video \url{https://youtu.be/RuGJjjVcV5w}.
\subsection{Experimental Setup}

The MIT humanoid is simulated using \textit{ode45} in MATLAB. The robot geometry is approximated with 4 spheres and 14 capsules in total, as shown by the green bodies in Fig. \ref{fig:with_without_collision}b. In practice, 15 relevant collision body pairs are considered in the CBF-WBC. The derivatives needed for the signed distance functions and CBFs are formulated using CasADi's algorithmic differentiation framework \cite{anderssonCasADiSoftwareFramework2019}. The quadratic programs for the CBF-WBC and LMPC are solved using qpSWIFT \cite{pandalaQpSWIFTRealTimeSparse2019} with an average solve time of around 0.2~ms on a Intel i9-11900K CPU.

\subsection{Hand Trajectory Tracking Experiment}
\begin{figure}[!htbp]
\centering
\includegraphics[scale=1]{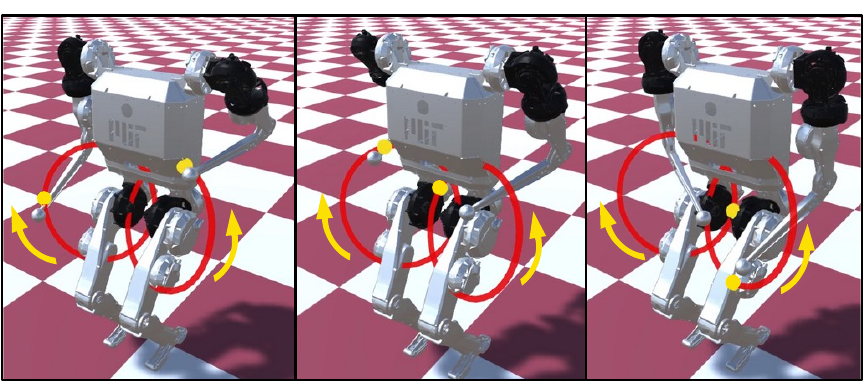}
\caption{Hand trajectory tracking experiment for an angular frequency of 1.57~rad/s. The reference trajectories shown in red are designed to penetrate the torso and legs. With CBFs, self-collisions are avoided. The yellow dots indicated the current reference position.}
\label{fig:arm_tracking}
\end{figure}

To illustrate the effectiveness of the CBF-WBC in avoiding self-collisions, the hands are commanded to track 3D circular trajectories that would nominally collide with the torso and legs (Fig. \ref{fig:arm_tracking}).
Only the CBF-WBC is used to balance the robot with a torso pose task while an arm end-effector task commands the positions of the hands.
Only signed distance CBF constraints \eqref{eq:coll-CBF-ineq} are used  for this experiment. As shown in Fig. \ref{fig:arm_tracking}, the CBF-WBC coordinates the arms to avoid self-collisions while tracking the reference point (shown as yellow dots in Fig. \ref{fig:arm_tracking}) as closely as possible.
\par
\begin{figure}[!htbp]
\centering
\includegraphics[scale=0.95]{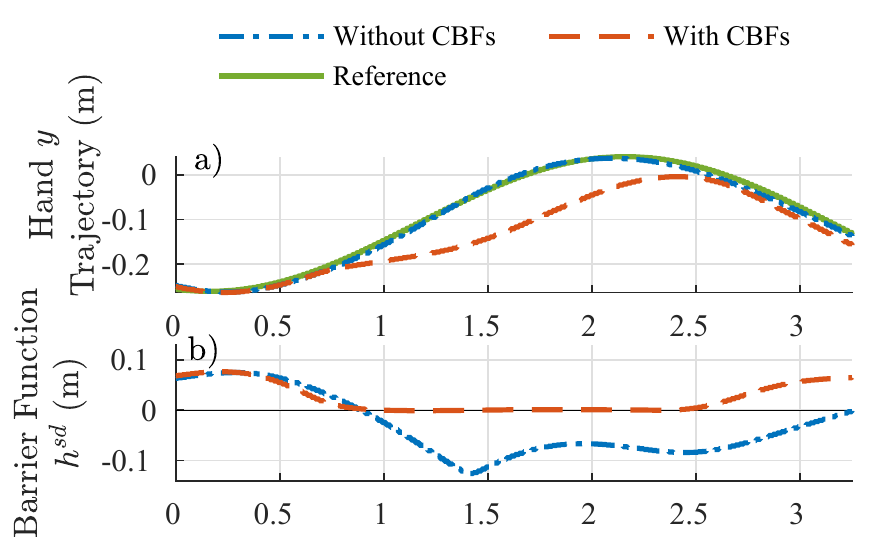}
\caption{Results from the hand trajectory tracking experiment. a) Tracking data and b) signed distance barrier function for the collision pair between the right forearm and the torso for an angular frequency of 1.57~rad/s.}
\label{fig:cbf_vs_no_cbf}
\end{figure}

The CBFs allow the robot to deviate from the reference trajectory to avoid self-collisions. Fig. \ref{fig:cbf_vs_no_cbf} compares the hand trajectory tracking in the $y$ direction (Fig. \ref{fig:cbf_vs_no_cbf}a) and the signed distance barrier function between the right forearm and the torso (Fig. \ref{fig:cbf_vs_no_cbf}b) with and without CBFs. Without CBFs, the tracking error is low throughout the experiment (Fig. \ref{fig:cbf_vs_no_cbf}a), but the forearm penetrates the torso by more than 0.1~m, as indicated by the negative signed distance barrier function (Fig. \ref{fig:cbf_vs_no_cbf}b). With the CBFs, the tracking error is initially low, but the robot must deviate from the reference trajectory at $\sim0.8$~s to prevent the collision. After $\sim2.5$~s, the risk of collision disappears, and the hand can follow the reference again.
\begin{figure}[!htbp]
\centering
\includegraphics[scale=1]{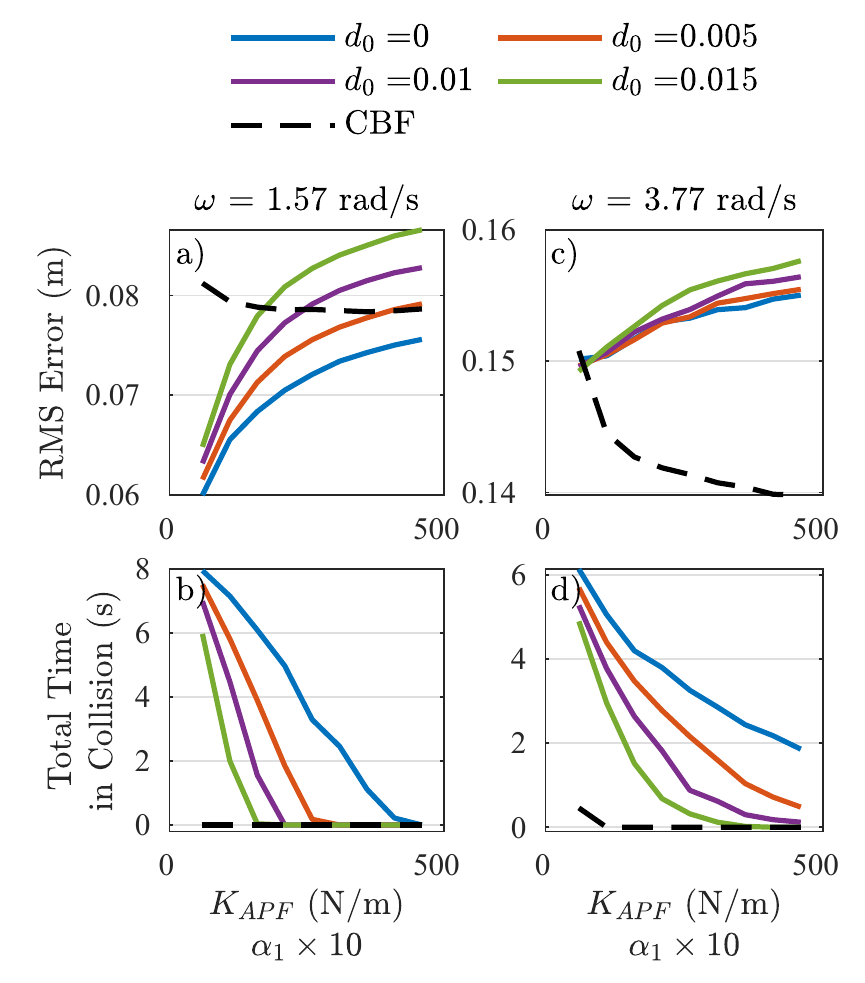}
\caption{Comparison of CBFs with APFs for angular frequencies $\omega=$ 1.57~rad/s (a,b) and $\omega=$3.77~rad/s (c,d). Top row : root mean square hand tracking error. Bottow row : total time spent in collision for all collision pairs.}
\label{fig:apf_vs_cbf}
\end{figure}
\par
The performance of CBF-WBC is benchmarked with artificial potential fields (APFs). APF-based collision avoidance tasks are implemented in the whole-body controller, where the signed distance functions $d_{AB} = h^\mathrm{sd}_{AB}(\bm{q})$ are used as task variables for all collisions pairs. The commanded task acceleration $\ddot{d}^\mathrm{cmd}_{AB}$ is defined as
\begin{gather}
\ddot{d}^\mathrm{cmd}_{AB}=\left\{
    \begin{array}{ll}
        K_\mathrm{APF}(d_0 - d_{AB})-2\sqrt{K_\mathrm{APF}}\dot{d}_{AB},& d_{AB} \le d_0  \\
        0 \qquad & d_{AB} > d_0,
    \end{array}
\right.
\end{gather}
where $d_0$ is a distance threshold below which the APF is active for the collision pair $AB$; $K_\mathrm{APF}$ is the stiffness of the potential field
\par
The performance of CBFs and APFs is compared for two angular frequencies $\omega \in \{1.57, 3.77\}$~rad/s. For these two frequencies, the hand trajectory tracking experiment is repeated multiple times with CBFs that use various values of $\alpha_1^{AB}=\alpha_2^{AB} \in \{6,11,...,46\}$~s$^{-1}$ and with APFs that use values of $(K_{APF}, d_{0}) \in \{60,110,...,460\}~\mathrm{N/m}\times\{0,0.005,...,0.015\}~\mathrm{m}$.
\par
\begin{figure*}[]
\centering
\includegraphics[scale=0.97]{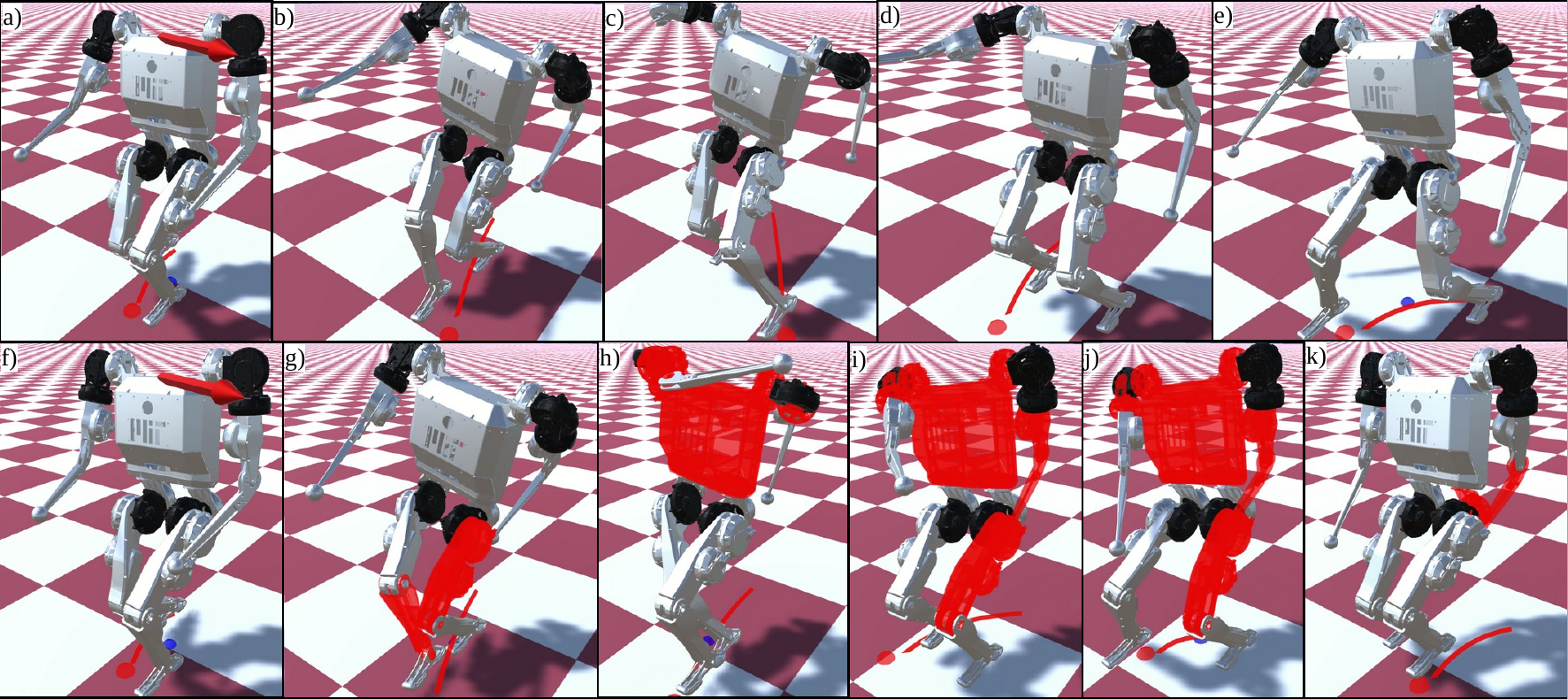}
\caption{Recovery behavior while walking at 1~m/s after a push of 25~N and 42~N in the forward and lateral directions, respectively. a-e) The CBFs ensure joint limit and self-collision avoidance. f-k) Without CBFs, many self-collisions occur.}
\label{fig:walking_animation}
\end{figure*}
Fig. \ref{fig:apf_vs_cbf} shows the root-mean-square (RMS) hand tracking error and the total time spent in collision for each simulation. For $\omega = 1.57$~rad/s, the RMS error for the CBFs rapidly reduces below $0.08$~m (Fig. \ref{fig:apf_vs_cbf}a) and all experiments remain collision-free (Fig. \ref{fig:apf_vs_cbf}b). For the APFs, many combinations of $(K_{APF}, d_0)$ result in collisions, but there are still values of $(K_{APF}, d_0)$ that are collision-free (Fig. \ref{fig:apf_vs_cbf}b) and that achieve the same tracking error as CBFs (Fig. \ref{fig:apf_vs_cbf}a) for this specific angular frequency ($\omega = 1.57$~rad/s).
\par

However, when increasing the angular frequency to $\omega = 3.77$~rad/s (Fig. \ref{fig:apf_vs_cbf}c-d), APFs do not match the performance of CBFs. CBFs achieve lower tracking error without collisions for $\alpha_1^{AB} \ge 11$, whereas the APFs-based controllers collide for most combinations of $(K_{APF},d_0)$ and induce higher tracking error as $K_{APF}$ increases.
\par
In all, while APFs can perform comparably to CBFs when specifically tailored for the task, CBFs are easier to tune since their performance is maintained across different angular frequencies and a single parameter has to be adjusted.

\subsection{Walking Experiments}
Walking experiments are performed to show the effectiveness of the CBF-WBC in avoiding joint limits and self-collisions for dynamic locomotion and disturbance recovery scenarios, where the risk of self-collisions is high. For these experiments, both the LMPC and the CBF-WBC are used. A torso pose task tracks the state predictions from the LMPC, and a foot pose task tracks a reference trajectory towards the predicted step location. A centroidal angular momentum task generates arm motions that assist locomotion and disturbance recovery, and a joint position task regularizes the arms of the robot to a nominal configuration. The signed distance CBF constraints \eqref{eq:coll-CBF-ineq} are used for all relevant collision pairs. The joint limit CBF constraints \eqref{eq:CBF_ineq_jmin} and \eqref{eq:CBF_ineq_jmax} are used for the knees and the arm joints.
Fig. \ref{fig:walking_animation}a-e illustrates the overall behavior of the combined LMPC and CBF-WBC while walking at 1~m/s and recovering from a push. Fig. \ref{fig:walking_animation}f-k illustrate the same case without CBFs, with self-collisions occurring between the arms, legs and torso.

\subsubsection{Swing Leg Self-Collision Avoidance}
Since the LMPC is not considering the full-body kinematics when choosing footstep locations, collisions between the legs can occur when the swing leg is moving towards the stance leg, as shown in Fig. \ref{fig:walking_animation}g. By incorporating signed distance CBF constraints \eqref{eq:coll-CBF-ineq}, CBF-WBC resolves kinematic feasibility conflicts when the desired footstep location induces a swing leg trajectory prone to self-collision (Fig. \ref{fig:walking_animation}b).
\par
\begin{figure}[!htbp]
\centering
\includegraphics[scale=1]{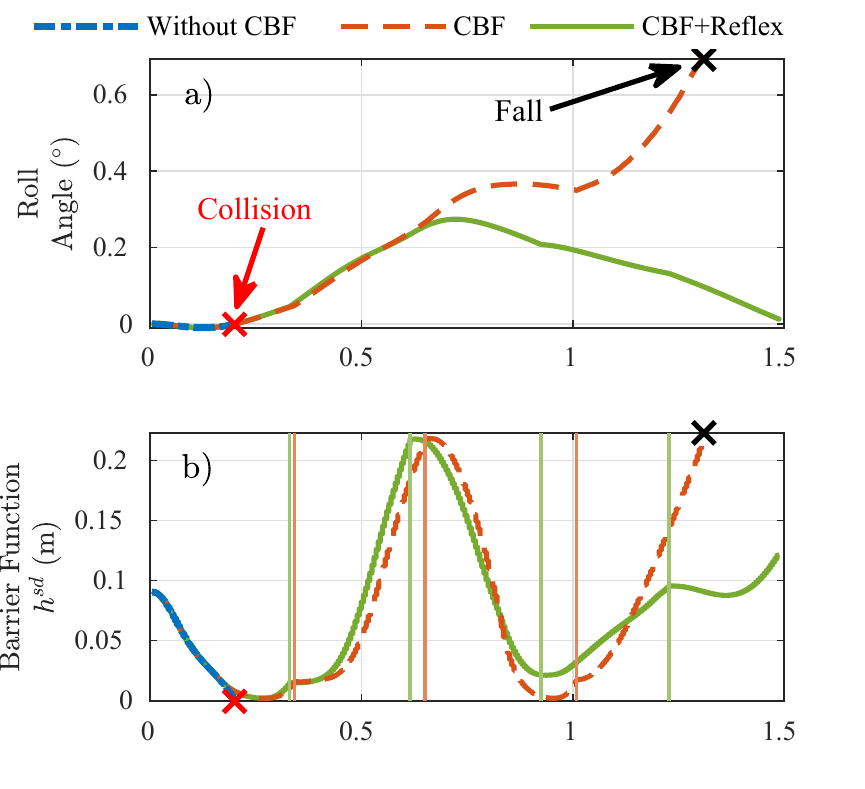}
\caption{Effect of the swing leg reflex in a 68~N lateral push scenario. a) Torso roll angle and b) the minimum distance between the shanks are shown. The vertical lines in b) show the footstep timing for each respective approach.}
\label{fig:cbf_reflex}
\end{figure}

However, since the CBF-WBC is purely reactive, the swing leg cannot always move around the stance leg depending on the convexity of the collision primitives. This can impede the controller from placing a step in time, causing the robot to fall (Fig. \ref{fig:cbf_reflex}). To address this issue, a heuristic-based swing leg reflex is designed to enhance the robustness of locomotion. According to this heuristic, the step frequency should increase as the collision risk increases. Let $d$ be the minimum distance between two legs. If $d$ drops below some threshold, the robot should initiate a new contact earlier to stabilize itself. To encode this behavior, the step period duration $T$ in the LMPC is adjusted as follows
\begin{equation}
T = \begin{cases}
    T_{\max} & d \geq d_{\max} \\
    \frac{T_{\max}-T_{\min}}{d_{\max}-d_{\min}}d + T_{\min} & d_{\min} < d < d_{\max} \\
    T_{\min} & d \leq d_{\min}
    \end{cases}
\end{equation}
where $T_{\max}$ and $T_{\min}$ are the nominal and minimum period time, respectively; $d_{\max}$ and $d_{\min}$ are the upper and lower bound on $d$, respectively.

Fig. \ref{fig:cbf_reflex} shows a case where the robot gets pushed 68~N for 0.2~s in the lateral direction. In order to recover, LMPC chooses a step location that coincides with the current stance foot's location. The WBC without CBFs tracks a naive swing trajectory which results in self-collision 0.2~s after the start of the disturbance. Without the swing reflex, the CBF-WBC avoids leg collisions, but the robot eventually falls. By adding the swing reflex, the robot recovers by increasing its step frequency (Fig \ref{fig:cbf_reflex}b) while remaining collision-free.

\subsubsection{Arm Self-Collision Avoidance}
\label{sec:Centroidal_Results}
As illustrated in Fig. \ref{fig:walking_animation}, the CBF-WBC generates arm motions that assist the robot's recovery behavior while respecting joint limits and avoiding self-collisions.
\par
To further investigate the benefits of the arm motions emerging from the CBF-WBC, the following strategies are compared in a scenario where the robot is pushed $29$~N laterally for $0.2$~s and is commanded to stop:
\par
\begin{itemize}
    \item No CBFs are used in the WBC
    \item Instead of CBFs, a joint position task regularizes the arms towards a feasible nominal configuration
    \item The CBF-WBC with reduced pose regularization
\end{itemize}
\par
\begin{figure}[!h]
\centering
\includegraphics[scale=1]{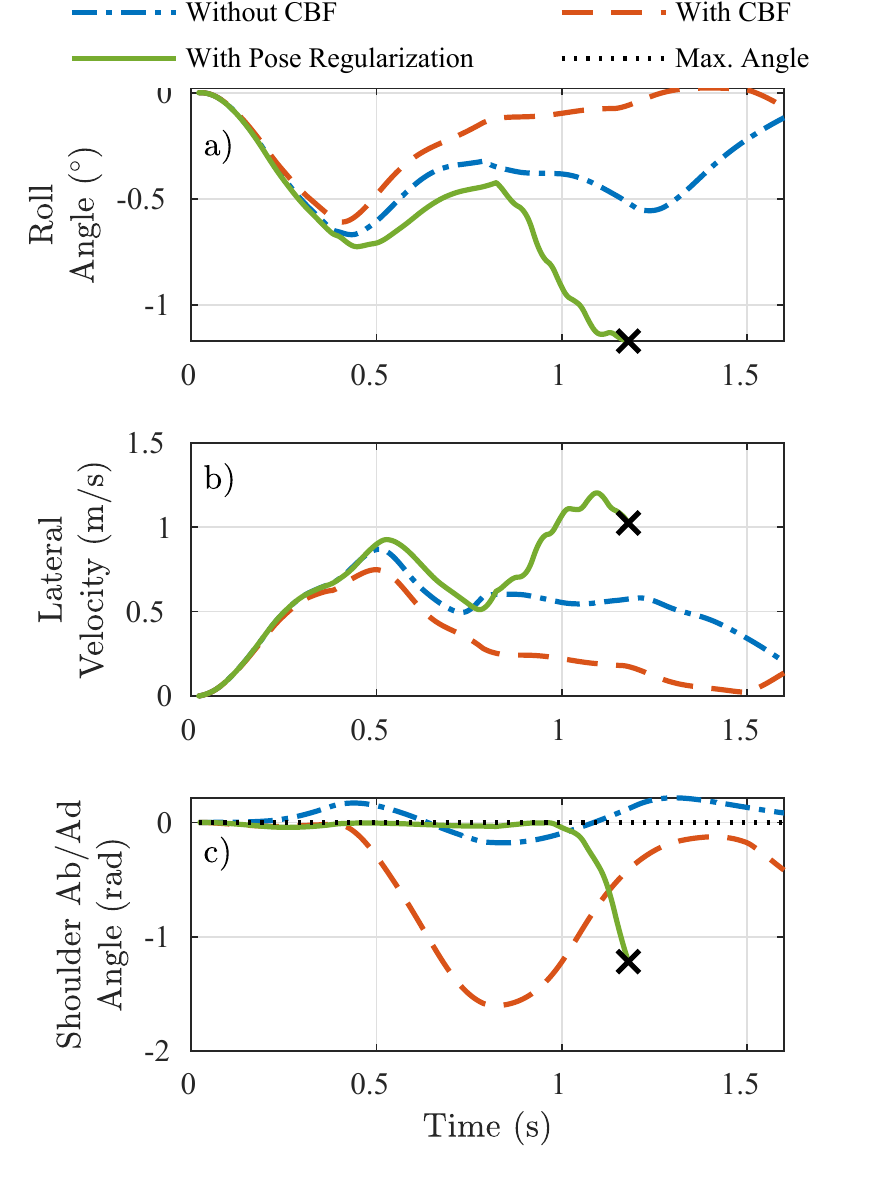}
\caption{Effect of pose regularization vs joint limit control barrier functions on the roll angle (a), lateral velocity (b), and shoulder abduction/adduction angle (c) after a lateral push of 29~N. The black crosses indicate a fall.}
\label{fig:3_cases_arms}
\end{figure}

Fig. \ref{fig:3_cases_arms} shows the torso roll angle, lateral velocity and right shoulder abduction-adduction angle as the robot undergoes the push for these three cases. In the first case without CBFs (Fig. \ref{fig:3_cases_arms}), the arms help the robot recover from the push, but they are penetrating the torso because the shoulder abduction/adduction joint limit is not enforced (Fig. \ref{fig:3_cases_arms}c). In the second case, pose regularization results in feasible arm motions but limits the effectiveness of the arms, causing the robot to fall. In the last case with CBFs, only a small amount of pose regularization is needed as the CBFs ensure kinematic feasibility. As a result, the range of motion of the shoulder abduction/adduction joint is higher, allowing the robot to exploit its arms better to avoid falling while respecting joint limits and avoiding self-collisions.

%% file: Content/Conclusion.tex
This paper develops a reactive controller that can respect joint limits and avoid self-collisions by incorporating control barrier functions in a QP-based whole-body controller that considers the full dynamics of the robot.
\par
This approach ensures collision-free trajectories for a simple hand-reaching experiment and dynamic locomotion. For the latter, CBFs constraints enable the WBC to avoid self-collisions when the robot's legs would otherwise collide and to generate feasible arm motions that assist locomotion.
\par
Nonetheless, the current strategy to avoid leg self-collisions is still limited. First, while CBFs ensure the kinematic feasibility of the swing trajectories, the CBFs can also prevent the legs from reaching the desired step location. Second, if the footstep planned by LMPC is truly infeasible, the LMPC planner should be informed accordingly. Future work will focus on developing a leg trajectory planner that can complement the CBF-WBC to generate feasible swing trajectories and inform the high-level planner of the infeasibility.

%% file: root.bbl
\begin{thebibliography}{10}
\providecommand{\url}[1]{#1}
\csname url@samestyle\endcsname
\providecommand{\newblock}{\relax}
\providecommand{\bibinfo}[2]{#2}
\providecommand{\BIBentrySTDinterwordspacing}{\spaceskip=0pt\relax}
\providecommand{\BIBentryALTinterwordstretchfactor}{4}
\providecommand{\BIBentryALTinterwordspacing}{\spaceskip=\fontdimen2\font plus
\BIBentryALTinterwordstretchfactor\fontdimen3\font minus
  \fontdimen4\font\relax}
\providecommand{\BIBforeignlanguage}[2]{{%
\expandafter\ifx\csname l@#1\endcsname\relax
\typeout{** WARNING: IEEEtran.bst: No hyphenation pattern has been}%
\typeout{** loaded for the language `#1'. Using the pattern for}%
\typeout{** the default language instead.}%
\else
\language=\csname l@#1\endcsname
\fi
#2}}
\providecommand{\BIBdecl}{\relax}
\BIBdecl

\bibitem{daiWholebodyMotionPlanning2014}
H.~Dai, A.~Valenzuela, and R.~Tedrake, ``Whole-body motion planning with
  centroidal dynamics and full kinematics,'' in \emph{2014 {{IEEE-RAS
  International Conference}} on {{Humanoid Robots}}}, Nov. 2014, pp. 295--302.

\bibitem{schulmanMotionPlanningSequential2014}
J.~Schulman, Y.~Duan, J.~Ho, A.~Lee, I.~Awwal, H.~Bradlow, J.~Pan, S.~Patil,
  K.~Goldberg, and P.~Abbeel, ``Motion planning with sequential convex
  optimization and convex collision checking,'' \emph{The International Journal
  of Robotics Research}, vol.~33, no.~9, pp. 1251--1270, Aug. 2014.

\bibitem{guReactiveLocomotionDecisionMaking2022}
Z.~Gu, N.~Boyd, and Y.~Zhao, ``Reactive {{Locomotion Decision-Making}} and
  {{Robust Motion Planning}} for {{Real-Time Perturbation Recovery}},'' Mar.
  2022.

\bibitem{dingDynamicWalkingFootstep2022}
Y.~Ding, C.~Khazoom, M.~Chignoli, and S.~Kim, ``Dynamic {{Walking}} with
  {{Footstep Adaptation}} on the {{MIT Humanoid}} via {{Linear Model Predictive
  Control}},'' Jun. 2022.

\bibitem{wensingHighspeedHumanoidRunning2013}
P.~M. Wensing and D.~E. Orin, ``High-speed humanoid running through control
  with a {{3D-SLIP}} model,'' in \emph{2013 {{IEEE}}/{{RSJ International
  Conference}} on {{Intelligent Robots}} and {{Systems}}}, Nov. 2013, pp.
  5134--5140.

\bibitem{sugiharaRealtimeHumanoidMotion2002}
T.~Sugihara, Y.~Nakamura, and H.~Inoue, ``Real-time humanoid motion generation
  through {{ZMP}} manipulation based on inverted pendulum control,'' in
  \emph{Proceedings 2002 {{IEEE International Conference}} on {{Robotics}} and
  {{Automation}} ({{Cat}}. {{No}}.{{02CH37292}})}, vol.~2, May 2002, pp.
  1404--1409 vol.2.

\bibitem{chiuCollisionFreeMPCWholeBody}
J.-R. Chiu, J.-P. Sleiman, M.~Mittal, F.~Farshidian, and M.~Hutter, ``A
  {{Collision-Free MPC}} for {{Whole-Body Dynamic Locomotion}} and
  {{Manipulation}},'' p.~8.

\bibitem{khatibRealtimeObstacleAvoidance1985}
O.~Khatib, ``Real-time obstacle avoidance for manipulators and mobile robots,''
  in \emph{1985 {{IEEE International Conference}} on {{Robotics}} and
  {{Automation Proceedings}}}, vol.~2, Mar. 1985, pp. 500--505.

\bibitem{schwienbacherSelfcollisionAvoidanceAngular2011}
M.~Schwienbacher, T.~Buschmann, S.~Lohmeier, V.~Favot, and H.~Ulbrich,
  ``Self-collision avoidance and angular momentum compensation for a biped
  humanoid robot,'' in \emph{2011 {{IEEE International Conference}} on
  {{Robotics}} and {{Automation}}}, May 2011, pp. 581--586.

\bibitem{dariushConstrainedResolvedAcceleration2010}
B.~Dariush, G.~Bin~Hammam, and D.~Orin, ``Constrained resolved acceleration
  control for humanoids,'' in \emph{2010 {{IEEE}}/{{RSJ International
  Conference}} on {{Intelligent Robots}} and {{Systems}}}, Oct. 2010, pp.
  710--717.

\bibitem{mirrazavisalehianUnifiedFrameworkCoordinated2018}
S.~S. Mirrazavi~Salehian, N.~Figueroa, and A.~Billard, ``A unified framework
  for coordinated multi-arm motion planning,'' \emph{The International Journal
  of Robotics Research}, vol.~37, no.~10, pp. 1205--1232, Sep. 2018.

\bibitem{koptevRealTimeSelfCollisionAvoidance2021}
M.~Koptev, N.~Figueroa, and A.~Billard, ``Real-{{Time Self-Collision
  Avoidance}} in {{Joint Space}} for {{Humanoid Robots}},'' \emph{IEEE Robotics
  and Automation Letters}, vol.~6, no.~2, pp. 1240--1247, Apr. 2021.

\bibitem{wensingImprovedComputationHumanoid2016}
P.~M. Wensing and D.~E. Orin, ``Improved {{Computation}} of the {{Humanoid
  Centroidal Dynamics}} and {{Application}} for {{Whole-Body Control}},''
  \emph{International Journal of Humanoid Robotics}, vol.~13, no.~01, p.
  1550039, Mar. 2016.

\bibitem{schullerOnlineCentroidalAngular2021}
R.~Schuller, G.~Mesesan, J.~Englsberger, J.~Lee, and C.~Ott, ``Online
  {{Centroidal Angular Momentum Reference Generation}} and {{Motion
  Optimization}} for {{Humanoid Push Recovery}},'' \emph{IEEE Robotics and
  Automation Letters}, vol.~6, no.~3, pp. 5689--5696, Jul. 2021.

\bibitem{grandiaMultiLayeredSafetyLegged2021}
R.~Grandia, A.~J. Taylor, A.~D. Ames, and M.~Hutter, ``Multi-{{Layered Safety}}
  for {{Legged Robots}} via {{Control Barrier Functions}} and {{Model
  Predictive Control}},'' Jun. 2021.

\bibitem{nguyen3DDynamicWalking2016}
Q.~Nguyen, A.~Hereid, J.~W. Grizzle, A.~D. Ames, and K.~Sreenath, ``{{3D}}
  dynamic walking on stepping stones with control barrier functions,'' in
  \emph{2016 {{IEEE}} 55th {{Conference}} on {{Decision}} and {{Control}}
  ({{CDC}})}, Dec. 2016, pp. 827--834.

\bibitem{singletaryComparativeAnalysisControl2020}
A.~Singletary, K.~Klingebiel, J.~Bourne, A.~Browning, P.~Tokumaru, and A.~Ames,
  ``Comparative {{Analysis}} of {{Control Barrier Functions}} and {{Artificial
  Potential Fields}} for {{Obstacle Avoidance}},'' Oct. 2020.

\bibitem{amesControlBarrierFunction2014a}
A.~D. Ames, J.~W. Grizzle, and P.~Tabuada, ``Control barrier function based
  quadratic programs with application to adaptive cruise control,'' in
  \emph{53rd {{IEEE Conference}} on {{Decision}} and {{Control}}}, Dec. 2014,
  pp. 6271--6278.

\bibitem{singletarySafetyCriticalManipulationCollisionFree2022}
A.~Singletary, W.~Guffey, T.~G. Molnar, R.~Sinnet, and A.~D. Ames,
  ``Safety-{{Critical Manipulation}} for {{Collision-Free Food Preparation}},''
  May 2022.

\bibitem{wensingOptimizationControlDynamic2014}
P.~M. Wensing, ``Optimization and {{Control}} of {{Dynamic Humanoid Running}}
  and {{Jumping}},'' Ph.D. dissertation, The Ohio State University, 2014.

\bibitem{kimHighlyDynamicQuadruped2019}
D.~Kim, J.~Di~Carlo, B.~Katz, G.~Bledt, and S.~Kim, ``Highly {{Dynamic
  Quadruped Locomotion}} via {{Whole-Body Impulse Control}} and {{Model
  Predictive Control}},'' \emph{arXiv:1909.06586 [cs]}, Sep. 2019.

\bibitem{vfi_Quiroz}
J.~J. Quiroz-Omaña and B.~V. Adorno, ``Whole-body control with (self)
  collision avoidance using vector field inequalities,'' \emph{IEEE Robotics
  and Automation Letters}, vol.~4, no.~4, pp. 4048--4053, 2019.

\bibitem{amesControlBarrierFunctions2019}
A.~D. Ames, S.~Coogan, M.~Egerstedt, G.~Notomista, K.~Sreenath, and P.~Tabuada,
  ``Control {{Barrier Functions}}: {{Theory}} and {{Applications}},'' Mar.
  2019.

\bibitem{nguyenExponentialControlBarrier2016}
Q.~Nguyen and K.~Sreenath, ``Exponential {{Control Barrier Functions}} for
  enforcing high relative-degree safety-critical constraints,'' in \emph{2016
  {{American Control Conference}} ({{ACC}})}, Jul. 2016, pp. 322--328.

\bibitem{lumelskyFastComputationDistance1985}
V.~J. Lumelsky, ``On fast computation of distance between line segments,''
  \emph{Information Processing Letters}, vol.~21, no.~2, pp. 55--61, Aug. 1985.

\bibitem{chignoliMITHumanoidRobot2021}
M.~Chignoli, D.~Kim, E.~{Stanger-Jones}, and S.~Kim, ``The {{MIT Humanoid
  Robot}}: {{Design}}, {{Motion Planning}}, and {{Control For Acrobatic
  Behaviors}},'' in \emph{2020 {{IEEE-RAS}} 20th {{International Conference}}
  on {{Humanoid Robots}} ({{Humanoids}})}, Jul. 2021, pp. 1--8.

\bibitem{anderssonCasADiSoftwareFramework2019}
J.~A.~E. Andersson, J.~Gillis, G.~Horn, J.~B. Rawlings, and M.~Diehl,
  ``{{CasADi}}: {{A}} software framework for nonlinear optimization and optimal
  control,'' \emph{Mathematical Programming Computation}, vol.~11, no.~1, pp.
  1--36, Mar. 2019.

\bibitem{pandalaQpSWIFTRealTimeSparse2019}
A.~G. Pandala, Y.~Ding, and H.-W. Park, ``{{qpSWIFT}}: {{A Real-Time Sparse
  Quadratic Program Solver}} for {{Robotic Applications}},'' \emph{IEEE
  Robotics and Automation Letters}, vol.~4, no.~4, pp. 3355--3362, Oct. 2019.

\end{thebibliography}
